\pdfoutput=1

\documentclass[11pt]{article}

\usepackage[]{acl}

\usepackage{times}
\usepackage{latexsym}
\usepackage{graphicx}
\usepackage[T1]{fontenc}

\usepackage[utf8]{inputenc}

\usepackage{microtype}
\usepackage[noorphans,vskip=0ex]{quoting}

\title{Combining Humor and Sarcasm for Improving Political Parody Detection}

  \author{
    {\bf Xiao Ao$^\alpha$} \quad {\bf Danae S\'{a}nchez Villegas$^\alpha$} \quad {\bf Daniel Preo\c{t}iuc-Pietro$^\beta$} \quad {\bf Nikolaos Aletras$^\alpha$}\\
    $^\alpha$ Computer Science Department, University of Sheffield, UK\\
    $^\beta$ Bloomberg\\
    {\small
    {\tt \{xao3,dsanchezvillegas1,n.aletras\}@sheffield.ac.uk}}\\
    {\small
    {\tt dpreotiucpie@bloomberg.net}}
}

\begin{document}
\maketitle

\begin{abstract}
Parody is a figurative device used for mimicking entities for comedic or critical purposes. Parody is intentionally humorous and often involves sarcasm. This paper explores jointly modelling these figurative tropes with the goal of improving performance of political parody detection in tweets.  To this end, we present a \emph{multi-encoder} model that combines three parallel encoders to enrich parody-specific representations with humor and sarcasm information. Experiments on a publicly available data set of political parody tweets demonstrate that our approach outperforms previous state-of-the-art methods.\footnote{Code is available here \url{https://github.com/iamoscar1/Multi_Encoder_Model_for_Political_Parody_Prediction}} 
\end{abstract}

\section{Introduction}

Parody is a figurative device which imitates entities such as politicians and celebrities by copying their particular style or a situation where the entity was involved \citep{rose1993parody}. It is an intrinsic part of social media as a relatively new comedic form \cite{TweetReporting}. A very popular type of parody is political parody, which is used to express political opposition and civic engagement \citep{davis2018seriously}.

One of the hallmarks of parody expression is the deployment of other figurative devices, such as humor and sarcasm, as emphasized on studies of parody in linguistics \citep{haiman1998talk,Parody_Humor}. For example, in Table \ref{tab:tweet} the text expresses sarcasm about Myspace\footnote{\url{https://myspace.com}} 
being a \lq winning technology\rq, while mocking the fact that three more popular social media sites were unavailable. This example also highlights the similarities between parody and real tweets, which may pose issues to misinformation classification systems~\cite{mu2020identifying}.

\renewcommand{\arraystretch}{1.0}
\begin{table}[!t]
    \footnotesize
    \centering
    \begin{tabular}{|m{1.5cm}|m{5cm}|}
        \hline
        \rowcolor[gray]{.7} \textbf{Twitter Handle} & 
        \texttt{@Queen\_UK} \\
        \hline
        \textbf{Parody tweet} &  Boris Johnson on the phone. Very smug that \colorbox{pink}{\#myspace} hasn't gone down. Says he's always backed \colorbox{pink}{winning technologies} \#whatsappdown \#instagramdown \#FacebookIsDown \\
        \hline
    \end{tabular}
    \caption{Example of a \emph{parody} tweet\protect\footnotemark\ by the Twitter handle @Queen\_UK. Humor and \colorbox{pink}{sarcasm} are expressed simultaneously.}
   \label{tab:tweet}
\end{table}

\footnotetext[\thefootnote]{\url{https://twitter.com/Queen\_UK/status/1445103605355323393?t=FGMNsMVFF\_G2tABYxFmkFw\&s=07}}

These figurative devices have so far been studied in isolation to parody.
Previous work on modeling humor in computational linguistics has focused on identifying jokes, i.e., short comedic passages that end with a hilarious line \citep{hetzron1991structure}, based on linguistic features \citep{1400851,purandare-litman-2006-humor,kiddon-brun-2011-thats} and deep learning techniques \citep{chen-soo-2018-humor,weller-seppi-2019-humor,annamoradnejad2020colbert}.
Similarly, computational approaches for modeling sarcasm (i.e., a form of verbal irony used to mock or convey content) in texts have been explored \citep{davidov-etal-2010-semi,gonzalez-ibanez-etal-2011-identifying,liebrecht-etal-2013-perfect,Rajadesingan2015SarcasmDO,ghosh-etal-2020-interpreting,ghosh-etal-2021-laughing}, including multi-modal utterances, i.e. texts, images, and videos \citep{cai-etal-2019-multi,castro-etal-2019-towards,oprea-magdy-2020-isarcasm}.
Recently, parody has been studied with natural language processing (NLP) methods by \citet{maronikolakis-etal-2020-analyzing} who introduced a data set of political parody accounts.
Their method for automatic recognition of posts shared by political parody accounts on Twitter is solely based on vanilla transformer models.

In this paper, we hypothesize that humor and sarcasm information could guide parody specific text encoders towards detecting nuances of figurative language. For this purpose, we propose a \emph{multi-encoder} model (\S \ref{sec:model}) consisting of three parallel encoders that are subsequently fused for parody classification. The first encoder learns parody specific information subsequently enhanced using the representations learned by a humor and sarcasm encoder respectively.

Our contributions are: (1) new state-of-the-art results on political parody detection in Twitter, consistently improving predictive performance over previous work by \citet{maronikolakis-etal-2020-analyzing}; and (2) insights on the limitations of neural models in capturing various linguistic characteristics of parody from extensive qualitative and quantitative analyses.



\section{Multi-Encoder Model for Political Parody Prediction}
\label{sec:model}

\citet{maronikolakis-etal-2020-analyzing} define political parody prediction as a binary classification task where a social media post $T$, consisting of a sequence of tokens $T=\{t_1,...,t_n\}$, is classified as real or parody. Real posts have been authored by actual politicians (e.g., \texttt{\@realDonaldTrump}) while parody posts come from their corresponding parody accounts (e.g., \texttt{\@realDonaldTrFan}).

\begin{figure}[t!]
\centering
\includegraphics[width=5.5cm]{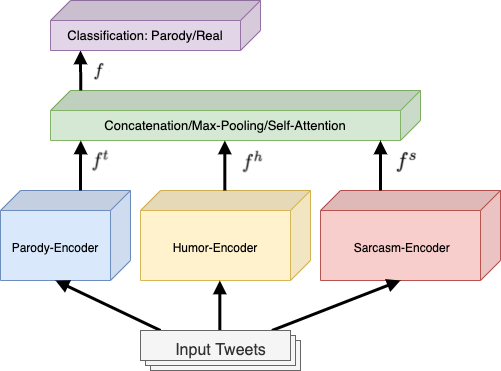}
\caption{The structure of our \emph{multi-encoder} model for combining humor and sarcasm information for political parody prediction.}
\label{fig:model}
\end{figure}

Parody tends to express complex tangled semantics of both humor and sarcasm simultaneously \citep{haiman1998talk,Parody_Humor}. To better exploit this characteristic of parody, we propose a \emph{multi-encoder} model that consists of three parallel encoders, a feature-fusion layer and a parody classification layer depicted in Fig.\ref{fig:model}.\footnote{Early experiments with multi-task learning did not result in improved performance. The results of these experiments can be found in Appendix \ref{appendix_MTL}.}.

\subsection{Text Encoders}

\paragraph{Parody}

As a task-specific parody encoder, we use the vanilla pretrained BERTweet \citep{nguyen-etal-2020-bertweet}, a BERT \cite{devlin-etal-2019-bert} based model pre-trained on a corpus of English Tweets and fine-tuned on the parody data set (\S \ref{sec:data}). 

\begin{figure}[t!]
\centering\includegraphics[width=7.5cm]{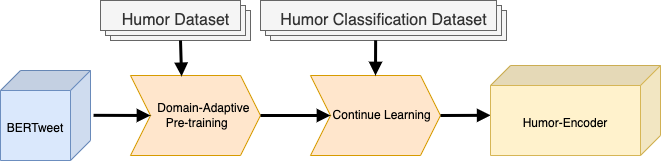}
\caption{\textit{Humor Encoder}.}
\label{fig:humorEncoder}
\end{figure}

\paragraph{Humor} 
To capture humor specific characteristics in social media text, we use the data set introduced by \citet{annamoradnejad2020colbert} which contains humorous and non-humorous short texts collected from Reddit and Huffington Post. First, we adapt BERTweet using domain-adaptive pre-training \citep{sun2020finetune,gururangan-etal-2020-dont} on 10,000 randomly selected humor-only short texts with masked language modeling. 
Subsequently, we use a continual learning strategy \citep{8107520,Sun_Wang_Li_Feng_Tian_Wu_Wang_2020} to gradually learn humor-specific properties by further fine-tuning BERTweet on a humor classification task (i.e., predicting whether a text is humorous or not) by using 40,000 randomly selected humorous and non-humorous short texts from the humor corpus described above (see Figure~\ref{fig:humorEncoder}).

\begin{figure}[t!]
\centering\includegraphics[width=7.5cm]{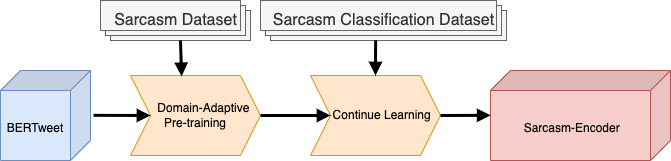}
\caption{\textit{Sarcasm Encoder}.}
\label{fig:sarcasmEncoder}
\end{figure}

\paragraph{Sarcasm} 
Similar to humor, we extract sarcasm-related semantic information from a post $T$ by using sarcasm annotated data sets from \citet{oprea-magdy-2020-isarcasm} and \citet{Rajadesingan2015SarcasmDO}. The first data set consists of 777 and 3,707 sarcasm and non-sarcasm posts from Twitter and the second data set consists of 9,104 sarcasm and more than 90,000 non-sarcasm posts from Twitter. We first perform domain-adaptive pre-training of BERTweet on all sarcastic posts with masked language modeling. Then, we fine-tune the model on a sarcasm classification task, similar to the humor encoder (see Figure~\ref{fig:sarcasmEncoder}). For the fine-tuning step, we use the 9,881 sarcastic tweets and 10,000 randomly sampled non-sarcasm tweets from the two data sets (i.e., 3,707 from the first and 6,293 from the second).



We compute parody $f^t$, humor $f^h$, and sarcasm $f^s$ representations by extracting the `classification' [CLS] token from each encoder respectively, where $f \in \mathbf{R}^{768}$.




\subsection{Combining Encoders}

We explore three approaches to combine $f^t$, $f^h$, and $f^s$ representations.

\paragraph{Concatenation} First, the three text representations are simply concatenated to form a combined representation $f \in \mathbf{R}^{768\times3}$.

\paragraph{Self-Attention} We also use a 4-head self-attention\footnote{Early experimentation with larger attention heads did not improve results in the dev set.} mechanism \cite{NIPS2017_3f5ee243} on $f^t, f^h, f^s$. The goal is to find correlations between representations and learn the contribution of each encoder in the final representation. 

\paragraph{Max-Pooling} Finally, we perform a max-pooling operation on each dimension of $f^t$, $f^h$, $f^s$ to obtain a representation $f \in \mathbf{R}^{768}$. The aim is to use the most dominant features learned by each encoder. 

\subsection{Classification}
Finally, we pass the combined representation $f$ to a classification layer with a sigmoid activation function for predicting whether a post is a parody or not. Three encoders are fine-tuned simultaneously on the parody data set (\S \ref{sec:data}).\footnote{Early experimentation with humor and sarcasm encoders frozen during the fine-tuning process did not show any performance improvement.}


\section{Experimental Setup}

\subsection{Data}
\label{sec:data}

We use the data set introduced by \citet{maronikolakis-etal-2020-analyzing} which contains 131,666 tweets written in English, with 65,956 tweets from political parody accounts and 65,710 tweets posted by real politician accounts. The data set is publicly available\footnote{\url{https://archive.org/details/parody_data_acl20}} and allows us to compare our results to state-of-the-art parody detection methods.  

We use the three data splits provided: (i) \emph{Person Split}, each split (train, dev, test) contains tweets from different real -- \emph{parody} account pairs; (ii) \emph{Gender Split}, two different splits based on the gender of the politicians (i.e., female accounts in train/dev and male in test, and male accounts in train/dev and female in test); \emph{Location Split}, data is split according to the location of the politicians in three groups (US, UK, Rest of the World or RoW). Each group is assigned to the test set and the other two groups to the train and dev sets.

\subsection{Baselines} 

We compare our \emph{multi-encoder} models with 
transformers for parody detection \citep{maronikolakis-etal-2020-analyzing}: {\bf BERT} \citep{devlin-etal-2019-bert} and {\bf RoBERTa} \citep{liu2019roberta}. Also, we compare our models to {\bf BERTweet} \citep{nguyen-etal-2020-bertweet}. 

\subsection{Implementation details}
\label{imp_details}

\paragraph{Humor Encoder} 
For adaptive pre-training, the batch-size is set to $16$ and the number of training epochs is set to $3$ with a learning rate of $2e^{-5}$. For humor classification, we use batch size of $128$ and the number of epochs is set to $2$ with a learning rate of $3e^{-5}$.

\paragraph{Sarcasm Encoder} 
We pretrain using a batch-size of $16$ over $5$ epochs with a learning rate of $2e^{-5}$. For fine-tuning on a sarcasm classification task, we use the $9,881$ sarcasm tweets and $10,000$ randomly sampled non-sarcasm tweets from the two data sets (i.e., $3,707$ from the first and $6,293$ from the second) using the same hyperparameters to the humor-specific encoder.

\paragraph{Multi-encoder}
For the complete \emph{multi-encoder} model, we use a batch size of $128$ and the learning rate is set to $2e^{-5}$. The entire model is fine-tuned for $2$ epochs.

\subsection{Evaluation} 
We evaluate the performance of all models using F1 score as \citet{maronikolakis-etal-2020-analyzing}. Results are obtained over $3$ runs using different random seeds reporting average and standard deviation. 


\section{Results}
\label{sec:results}

\subsection{Predictive Performance}
Table \ref{tab:person} shows the results for parody detection on the \emph{Person Split}. 
We observe that \textit{BERTweet} has the best performance (F1: $90.72$) among transformer-based models (\textit{BERT}, \textit{RoBERTa}, \textit{BERTweet}), outperforming previous state-of-the-art by \citet{maronikolakis-etal-2020-analyzing}. This is due to the fact that \textit{BERTweet} has been specifically pre-trained on Twitter text. Similar behavior is observed on the \emph{Gender} and \emph{Location} splits (see Table~\ref{tab:gender} and \ref{tab:location} respectively). 

\renewcommand{\arraystretch}{1.1}
\begin{table}[!t]
\small
\centering

\begin{tabular}{|l| c|}
\rowcolor[gray]{.7}\multicolumn{2}{|c|}{\textbf{Person}} \\\hline
\rowcolor[gray]{.7}\multicolumn{1}{|c|}{\textbf{Model}}& \multicolumn{1}{|c|}{\textbf{F1}} 
\\ \hline
\rowcolor[gray]{0.9}\multicolumn{2}{|l|}{\textbf{Single-Encoder}}\\
\cellcolor[gray]{1}BERT$^{\ast\ast}$ &  $87.65\pm0.18$ 
\\
\cellcolor[gray]{1}RoBERTa$^{\ast\ast}$ & $89.66\pm0.33$ 
\\ 
\cellcolor[gray]{1}BERTweet & $90.72\pm0.31$ 
\\\hline
\rowcolor[gray]{0.9} \multicolumn{2}{|l|}{\textbf{Multi-encoder (Ours)}}\\
\cellcolor[gray]{1}Concatenation & $88.99\pm0.17$ 
\\ 
\cellcolor[gray]{1}Self-Attention & $\textbf{91.19} \pm\textbf{0.31}$ 
\\ 
\cellcolor[gray]{1}Max-Pooling & $91.05\pm0.30$
\\ \hline
\end{tabular}
\caption{F1-scores for parody detection on the \emph{Person Split}. $^{\ast\ast}$ Results from \citet{maronikolakis-etal-2020-analyzing}. 
Best results are in bold.}
\label{tab:person}
\end{table}

\begin{table}[!t]
\renewcommand{\arraystretch}{1.1}
\centering
\small
\begin{tabular}{ |l| c|c|}
\hline
\rowcolor[gray]{.7}\multicolumn{3}{|c|}{\textbf{Gender}} \\\hline
\rowcolor[gray]{.7}\multicolumn{1}{|c|}{\textbf{Model}} & \textbf{M$\to$F} & \textbf{F$\to$M} \\ \hline
\rowcolor[gray]{0.9} \multicolumn{3}{|l|}{\textbf{Single-Encoder}}\\
\cellcolor[gray]{1}BERT$^{\ast\ast}$ & $85.85\pm0.28$ & $84.40\pm0.35$\\
\cellcolor[gray]{1}RoBERTa$^{\ast\ast}$ & $87.11\pm0.31$ & $84.87\pm0.38$\\
\cellcolor[gray]{1}BERTweet & $88.01\pm0.29$ &  $85.57\pm0.27$\\ \hline
\rowcolor[gray]{0.9}\multicolumn{3}{|l|}{\textbf{Multi-encoder (Ours)}}  \\
\cellcolor[gray]{1}Concatenation & $86.84\pm0.15$ & $84.21\pm0.22$\\
\cellcolor[gray]{1}Self-Attention & $\textbf{89.97}\pm\textbf{0.34}$ & $\textbf{88.56}\pm\textbf{0.39}$\\ 
\cellcolor[gray]{1}Max-Pooling & $88.39\pm0.27$ & $86.89\pm0.56$\\ \hline

\end{tabular}
\caption{F1-scores on the \emph{Gender Split}. 
$^{\ast\ast}$ Results from \citet{maronikolakis-etal-2020-analyzing}. Best results are in bold.}
\label{tab:gender}
\end{table}

Our proposed \emph{multi-encoder} achieves the best performance when using \textit{Self-Attention} to combine the three parallel encoders (F1: $91.19$; $89.97$, $88.56$; $88.37$, $87.91$, $87.16$; for \emph{Person}, \emph{Gender}, and \emph{Location} splits respectively). Moreover, it outperforms the best single-encoder model \textit{BERTweet} in the majority of cases which corroborates that parody detection benefits from combining general contextual representations with humor and sarcasm specific information, as humor and sarcasm are important characteristics of parody \citep{haiman1998talk,Parody_Humor}. On the other hand, simply concatenating the three parallel encoders degrades the performance across different splits (\emph{Person}: $88.99$; \emph{Gender}: $86.84$, $84.21$ \emph{Location}: $85.41$, $84.74$, $83.62$). This happens because the concatenation operation treats the three encoders as equally important. While humor and sarcasm are related to parody, they may not necessarily have the same relevance as indicators of parody. 

Our best performing model (\textit{Self-Attention}) outperforms the vanilla \textit{BERTweet} by $3$ F1 points when trained on female accounts and by almost $2$ F1 points when trained on male accounts. We speculate that the additional linguistic information from the two encoders (i.e., sarcasm and humor) is more beneficial in low data settings. The number of female politicians is considerably smaller than males in the data set (see \citet{maronikolakis-etal-2020-analyzing} for more details).



\renewcommand{\arraystretch}{1.1}
\begin{table}[!t]
\centering
\small
\resizebox{\linewidth}{!}{
\begin{tabular}{ |l|c|c|c|}
\hline
\rowcolor[gray]{.7}\multicolumn{4}{|c|}{\textbf{Location}}\\\hline
\rowcolor[gray]{.7}\multicolumn{1}{|c|}{\textbf{Model}} & \begin{tabular}{c}\textbf{UK+US}\\ \textbf{$\to$ RoW}\end{tabular}  & \begin{tabular}{c}\textbf{RoW+US} \\ \textbf{$\to$ UK} \end{tabular} 
& \begin{tabular}{c}\textbf{RoW+UK} \\ \textbf{$\to$ US}\end{tabular}\\ \hline 
\rowcolor[gray]{0.9} \multicolumn{4}{|l|}{\textbf{Single-Encoder}}\\
\cellcolor[gray]{1}BERT$^{\ast\ast}$ & $86.69\pm0.45$ & $83.78\pm0.19$ & $83.12\pm0.60$\\ 
\cellcolor[gray]{1}RoBERTa$^{\ast\ast}$  & $87.70\pm0.45$ & $85.10\pm0.27$ & $85.99\pm0.61$\\
\cellcolor[gray]{1}BERTweet & $88.21\pm0.26$ & $87.85\pm0.24$ & $\textbf{87.18}\pm\textbf{0.41}$\\ \hline
\rowcolor[gray]{0.9} \multicolumn{4}{|l|}{\textbf{Multi-encoder (Ours)}}\\
\cellcolor[gray]{1}Concatenation & $85.41\pm0.26$ & $84.74\pm0.20$ & $83.62\pm0.35$\\ 
\cellcolor[gray]{1}Self-Attention & $\textbf{88.37}\pm\textbf{0.28}$ & $\textbf{87.91}\pm\textbf{0.19}$ & $87.16\pm0.37$\\ 
\cellcolor[gray]{1}Max-Pooling & $88.25\pm0.39$ & $86.49\pm0.33$ & $86.54\pm0.41$\\ \hline

\end{tabular}}
\caption{F1-scores on the \emph{Location Split}. 
$^{\ast\ast}$ Results from \citet{maronikolakis-etal-2020-analyzing}. Best results are in bold.
}
\label{tab:location}
\end{table}

\subsection{Ablation Study} 
We also examine the effect of combining parody-specific representations with humor and sarcasm information by running an ablation study. We compare performance of four models: using parody representations only (P), and combining parody representations with humor (P+H), or sarcasm (P+S) information, as well as with both (P+S+H). The results of this analysis are depicted in Tables~\ref{tab:person_complete}, \ref{tab:gender_complete} and \ref{tab:location_complete}. We observe that both sarcasm and humor contribute to the performance gain, but using both is more beneficial. Modelling sarcasm leads to more gains than humor and this could be attributed to the characteristics of the parody corpus, namely that it focuses primarily on the political domain, which have a high sarcastic component \citep{anderson2017social}.



\renewcommand{\arraystretch}{1.2}
\begin{table}[!h]
\small
\centering
\begin{tabular}{|l| c|}
\rowcolor[gray]{.7}\multicolumn{2}{|c|}{\textbf{Person}} \\\hline
\rowcolor[gray]{.7}\multicolumn{1}{|c|}{\textbf{Model}}& \multicolumn{1}{|c|}{\textbf{F1}} 
\\ \hline
\rowcolor[gray]{0.9}\multicolumn{2}{|l|}{\textbf{Single-Encoder}}\\
\cellcolor[gray]{1}BERTweet (P) & $90.72\pm0.31$ 
\\\hline
\rowcolor[gray]{0.9} \multicolumn{2}{|l|}{\textbf{Multi-encoder (Ours)}}\\
\cellcolor[gray]{1}Concatenation (P+S+H) & $88.99\pm0.17$ 
\\ 
\cellcolor[gray]{1}Concatenation (P+S) & $90.51\pm0.26$ 
\\ 
\cellcolor[gray]{1}Concatenation (P+H) & $89.98\pm0.23$ 
\\ 
\cellcolor[gray]{1}Self-Attention (P+S+H) & $\textbf{91.19} \pm\textbf{0.31}$ 
\\ 
\cellcolor[gray]{1}Self-Attention (P+S) & $91.14\pm0.40$ 
\\ 
\cellcolor[gray]{1}Self-Attention (P+H)& $90.98\pm0.36$ 
\\ 
\cellcolor[gray]{1}Max-Pooling\hspace{0.44em} (P+S+H) & $91.05\pm0.30$
\\ 
\cellcolor[gray]{1}Max-Pooling\hspace{0.44em} (P+S) & $91.06\pm0.39$
\\ 
\cellcolor[gray]{1}Max-Pooling\hspace{0.44em} (P+H) & $90.78\pm0.42$
\\ \hline
\end{tabular}
\caption{F1-scores for parody detection on the \emph{Person Split} with various settings: parody (P) representations only, and combining parody representations with humor (P+H), or sarcasm (P+S) information, as well as with both (P+S+H). Best results are in bold.
}
\label{tab:person_complete}
\end{table}

\renewcommand{\arraystretch}{1.2}
\begin{table}[!t]
\centering
\small
\begin{tabular}{ |l| c|c|}
\hline
\rowcolor[gray]{.7}\multicolumn{3}{|c|}{\textbf{Gender}} \\\hline
\rowcolor[gray]{.7}\multicolumn{1}{|c|}{\textbf{Model}} & \textbf{M$\to$F} & \textbf{F$\to$M} \\ \hline
\rowcolor[gray]{0.9} \multicolumn{3}{|l|}{\textbf{Single-Encoder}}\\
\cellcolor[gray]{1}BERTweet (P) & $88.01\pm0.29$ &  $85.57\pm0.27$\\ \hline
\rowcolor[gray]{0.9}\multicolumn{3}{|l|}{\textbf{Multi-encoder (Ours)}}  \\
\cellcolor[gray]{1}Concatenation (P+S+H)& $86.84\pm0.15$ & $84.21\pm0.22$\\
\cellcolor[gray]{1}Concatenation (P+S)& $86.93\pm0.40$ & $83.70\pm0.41$\\
\cellcolor[gray]{1}Concatenation (P+H)& $86.58\pm0.31$ & $83.34\pm0.38$\\
\cellcolor[gray]{1}Self-Attention (P+S+H)& $\textbf{89.97}\pm\textbf{0.34}$ & $\textbf{88.56}\pm\textbf{0.39}$\\ 
\cellcolor[gray]{1}Self-Attention (P+S)& $89.49\pm0.37$ & $88.23\pm0.44$\\ 
\cellcolor[gray]{1}Self-Attention (P+H)& $88.71\pm0.42$ & $87.62\pm0.50$\\ 
\cellcolor[gray]{1}Max-Pooling\hspace{0.44em} (P+S+H)& $88.39\pm0.27$ & $86.89\pm0.56$\\ 
\cellcolor[gray]{1}Max-Pooling\hspace{0.44em} (P+S)& $88.36\pm0.46$ & $86.55\pm0.49$\\
\cellcolor[gray]{1}Max-Pooling\hspace{0.44em} (P+H)& $88.14\pm0.52$ & $86.53\pm0.53$\\\hline

\end{tabular}
\caption{
F1-scores for parody detection on the \emph{Gender Split} with various settings: parody (P) representations only, and combining parody representations with humor (P+H), or sarcasm (P+S) information, as well as with both (P+S+H). Best results are in bold.}
\label{tab:gender_complete}
\end{table}

\renewcommand{\arraystretch}{1.2}
\begin{table}[!t]
\centering
\small
\resizebox{\linewidth}{!}{
\begin{tabular}{ |l|c|c|c|}
\hline
\rowcolor[gray]{.7}\multicolumn{4}{|c|}{\textbf{Location}}\\\hline
\rowcolor[gray]{.7}\multicolumn{1}{|c|}{\textbf{Model}} & \begin{tabular}{c}\textbf{UK+US}\\ \textbf{$\to$ RoW}\end{tabular}  & \begin{tabular}{c}\textbf{RoW+US} \\ \textbf{$\to$ UK} \end{tabular} 
& \begin{tabular}{c}\textbf{RoW+UK} \\ \textbf{$\to$ US}\end{tabular}\\ \hline 
\rowcolor[gray]{0.9} \multicolumn{4}{|l|}{\textbf{Single-Encoder}}\\
\cellcolor[gray]{1}BERTweet (P) & $88.21\pm0.26$ & $87.85\pm0.24$ & $\textbf{87.18}\pm\textbf{0.41}$\\ \hline


\rowcolor[gray]{0.9} \multicolumn{4}{|l|}{\textbf{Multi-encoder (Ours)}}\\
\cellcolor[gray]{1}Concatenation (P+S+H) & $85.41\pm0.26$ & $84.74\pm0.20$ & $83.62\pm0.35$\\
\cellcolor[gray]{1}Concatenation (P+S) & $85.92\pm0.24$ & $85.67\pm0.18$ & $84.09\pm0.39$\\
\cellcolor[gray]{1}Concatenation (P+H) & $85.39\pm0.29$ & $85.33\pm0.26$ & $83.75\pm0.44$\\
\cellcolor[gray]{1}Self-Attention (P+S+H) & $\textbf{88.37}\pm\textbf{0.28}$ & $\textbf{87.91}\pm\textbf{0.19}$ & $87.16\pm0.37$\\ 
\cellcolor[gray]{1}Self-Attention (P+S) & $88.24\pm0.33$ & $87.88\pm0.23$ & $86.47\pm0.32$\\
\cellcolor[gray]{1}Self-Attention (P+H) & $88.13\pm0.35$ & $87.05\pm0.28$ & $85.36\pm0.40$\\
\cellcolor[gray]{1}Max-Pooling\hspace{0.44em}  (P+S+H) & $88.25\pm0.39$ & $86.49\pm0.33$ & $86.54\pm0.41$\\
\cellcolor[gray]{1}Max-Pooling\hspace{0.44em} (P+S) & $88.28\pm0.42$ & $87.83\pm0.39$ & $86.56\pm0.36$\\
\cellcolor[gray]{1}Max-Pooling\hspace{0.44em} (P+H) & $88.22\pm0.52$ & $86.44\pm0.42$ & $85.96\pm0.45$\\
\hline

\end{tabular}
}
\caption{
F1-scores for parody detection on the \emph{Location Split} with various settings: parody (P) representations only, and combining parody representations with humor (P+H), or sarcasm (P+S) information, as well as with both (P+S+H). Best results are in bold.}
\label{tab:location_complete}
\end{table}
\section{Error Analysis}

Finally, we perform an error analysis to examine the behavior and limitations of our best-performing model (\emph{multi-encoder} with Self-Attention). 

The next two examples correspond to real tweets that were misclassified as parody:
\begin{itemize}
\item [(1)]  \textit{Congratulations, <mention>! <url>.}
\item [(2)] \textit{It's a shame that Boris isn't here answering questions from the public this evening.}
\end{itemize}
We speculate that the model misclassified these tweets as parody because they contain terms that are related to sarcastic short texts such as user mentions, punctuation marks (\textit{!}), and negation (\textit{isn't}) \cite{gonzalez-ibanez-etal-2011-identifying,Parody_Humor}.

The following two examples correspond to parody tweets that were misclassified as real:
\begin{itemize}
\item [(3)] \textit{Hey America, it’s time to use your safe word.}
\item [(4)]\textit{I fully support the Digital Singles Market.}
\end{itemize}

Example (3) is a call-to-action message, while Example (4) is a statement expressing support for a particular subject. These statements are written in a style that is similar to political slogans or campaign speeches \citep{fowler2021political} that the model fails to recognise. As a result, in addition to humor and sarcasm semantics, the model might be improved by integrating knowledge from the political domain such as from political speeches.

\section{Conclusion}

In this paper, we studied the impact of jointly modelling figurative devices to improve predictive performance of political parody detection in tweets. Our motivation was based on  studies in linguistics which emphasize the humorous and sarcastic components of parody \citep{haiman1998talk,Parody_Humor}. We presented a method that combines parallel encoders to capture parody, humor, and sarcasm specific representations from input sequences, which outperforms previous state-of-the-art proposed by \citet{maronikolakis-etal-2020-analyzing}. 

In the future, we plan to combine information from other modalities (e.g., images) for improving parody detection~\citep{sanchez-villegas-aletras-2021-point,sanchez-villegas-etal-2021-analyzing}.

\section*{Acknowledgements}
We would like to thank all reviewers for their valuable feedback. DSV is supported by the Centre for Doctoral Training in Speech and Language Technologies (SLT) and their Applications funded by the
UK Research and Innovation grant EP/S023062/1.

\bibliography{anthology,custom}
\bibliographystyle{acl_natbib}

\appendix
\clearpage

\section{Multitask-Learning}
\label{appendix_MTL}
We also tested applying \emph{multi-task learning} approaches \citep{Caruana93multitasklearning:} to use either sarcasm prediction (P+S), humor prediction (P+H) or both (P+S+H) as auxiliary tasks for parody detection. We utilize BERTweet as the share encoder and independent classification layers for parody and humor or sarcasm. 
Three sets of weights are applied to losses from each independent classification layer and the three layers are stacked.
The best results are chosen and depicted in Table~\ref{tab:person_mtl}, Table \ref{tab:gender_mtl} and Table \ref{tab:location_mtl}.

\begin{table}[!h]
\small
\centering

\begin{tabular}{|l| c|}
\rowcolor[gray]{.7}\multicolumn{2}{|c|}{\textbf{Person}} \\\hline
\rowcolor[gray]{.7}\multicolumn{1}{|c|}{\textbf{Model}}& \multicolumn{1}{|c|}{\textbf{F1}} 
\\ \hline
\rowcolor[gray]{0.9}\multicolumn{2}{|l|}{\textbf{Single-Encoder}}\\
\cellcolor[gray]{1}BERTweet (P) & $\textbf{90.72}\pm\textbf{0.31}$ 
\\\hline
\rowcolor[gray]{0.9}\multicolumn{2}{|l|}{\textbf{Multi-Task}}\\
\cellcolor[gray]{1}P+S+H &  $87.46\pm0.18$ 
\\
\cellcolor[gray]{1}P+S & $89.41 \pm 0.31$ 
\\ 
\cellcolor[gray]{1}P+H & $87.41\pm0.38$ 
\\ \hline
\end{tabular}
\caption{F1-scores for parody detection on the \emph{Person Split} using Multi-task Learning models (P: Parody, S: Sarcasm, H: Humor). Best results are in bold.}
\label{tab:person_mtl}
\end{table}

\begin{table}[!h]
\renewcommand{\arraystretch}{1.1}
\centering
\small
\begin{tabular}{ |l| c|c|}
\hline
\rowcolor[gray]{.7}\multicolumn{3}{|c|}{\textbf{Gender}} \\\hline
\rowcolor[gray]{.7}\multicolumn{1}{|c|}{\textbf{Model}} & \textbf{M$\to$F} & \textbf{F$\to$M} \\ 
\rowcolor[gray]{0.9} \multicolumn{3}{|l|}{\textbf{Single-Encoder}}\\
\cellcolor[gray]{1}BERTweet (P) & $88.01\pm0.29$ &  $85.57\pm0.27$\\ \hline
\rowcolor[gray]{0.9} \multicolumn{3}{|l|}{\textbf{Multi-Task}}\\
\cellcolor[gray]{1}P+S+H & $85.28\pm0.29$ & $84.10\pm0.37$\\
\cellcolor[gray]{1}P+S & $\textbf{88.13}\pm \textbf{0.21}$ & $\textbf{86.07}\pm\textbf{0.44}$\\
\cellcolor[gray]{1}P+H & $84.53\pm0.31$ & $86.07\pm0.47$\\ \hline
\end{tabular}
\caption{F1-scores on the \emph{Gender Split} using Multi-task Learning models (P: Parody, S: Sarcasm, H: Humor). Best results are in bold.}
\label{tab:gender_mtl}
\end{table}

\begin{table}[!h]
\centering
\small
\resizebox{\linewidth}{!}{
\begin{tabular}{ |l|c|c|c|}
\hline
\rowcolor[gray]{.7}\multicolumn{4}{|c|}{\textbf{Location}}\\\hline
\rowcolor[gray]{.7}\multicolumn{1}{|c|}{\textbf{Model}} & \begin{tabular}{c}\textbf{UK+US}\\ \textbf{$\to$ RoW}\end{tabular}  & \begin{tabular}{c}\textbf{RoW+US} \\ \textbf{$\to$ UK} \end{tabular} 
& \begin{tabular}{c}\textbf{RoW+UK} \\ \textbf{$\to$ US}\end{tabular}\\ \hline 
\rowcolor[gray]{0.9} \multicolumn{4}{|l|}{\textbf{Single-Encoder}}\\
\cellcolor[gray]{1}BERTweet (P) & $\textbf{88.21}\pm\textbf{0.26}$ & $\textbf{87.85}\pm\textbf{0.24}$ & $\textbf{87.18}\pm\textbf{0.41}$\\ \hline
\rowcolor[gray]{0.9} \multicolumn{4}{|l|}{\textbf{Multi-Task}}\\
\cellcolor[gray]{1}P+S+H & $86.41\pm0.17$ & $86.23\pm0.20$ & $85.13\pm0.29$\\ 
\cellcolor[gray]{1}P+S & $87.74\pm0.36$ & $87.26\pm0.34$ & $86.67\pm0.43$\\
\cellcolor[gray]{1}P+H & $85.54\pm0.38$ & $84.78\pm0.47$ & $84.15\pm0.56$\\  \hline

\end{tabular}}
\caption{F1-scores on the \emph{Location Split} using Multi-task Learning models (P: Parody, S: Sarcasm, H: Humor). Best results are in bold.
}
\label{tab:location_mtl}
\end{table}

\end{document}